# Vegetation Stratum Occupancy Prediction from Airborne LiDAR 3D Point Clouds

E. Kalinicheva[1,2], L. Landrieu[1], C. Mallet[1], N. Chehata[1,3]

[1]Université Gustave Eiffel, IGN-ENSG, LASTIG, F-94160 Saint-Mandé, France
Email: {ekaterina.kalinicheva, loic.landrieu, clement.mallet}@ign.fr

[2]INRAE, UMR 1202 BIOGECO, Université de Bordeaux, France

[3]EA G&E Bordeaux INP, Université Bordeaux Montaigne, France
Email: nesrine.chehata@bordeaux-inp.fr

## 1. Introduction

Estimating the structure of vegetation is a crucial first step for many environmental and ecological applications (Daubenmire 1956). In particular, pasture land management requires estimating the occupancy of the different vegetation strata within agricultural parcels. This is a time-consuming undertaking, often performed with *in situ* ocular approximate measurements. Nowadays, airborne platforms allow public and private actors to gather high accuracy geometric and radiometric data over large areas (Chen 2007). Bolstered by the compelling improvements (Guo et al., 2020) and increased accessibility of deep learning for 3D point clouds, we propose a 3D deep learning method to estimate the occupancy of different vegetation strata from airborne LiDAR and camera sensors.

Our method predicts raster occupancy maps for three vegetation strata (lower, medium, and higher) from 3D point clouds. Our training scheme allows our network to only be supervised with aggregated occupancy values at the plot level, which are easier to produce than point or pixel-level annotations. We also propose to use priors on the stratum elevation and the occupancy maps to improve the legibility and interpretability of the resulting maps.

## 2. Data and Methods

We present a new open-access dataset allowing for training and evaluating stratum occupancy regression methods from 3D LiDAR data. We then propose our network architecture along a training scheme for inferring raster occupancy maps while only training with aggregated values.

### 2.1 Dataset Composition

Our proposed dataset comprises 199 cylindrical plots of 10 m radius corresponding to typical pasture land parcels in South-Eastern France. Each plot contains between 3000 and 17000 3D points, and each point is attributed with a total of 9 features: (i) absolute 3D coordinates, (ii) RGB and Near-InfraRed reflectance obtained with aerial cameras, (iii) uncalibrated laser intensity and return number provided by the aerial LiDAR.

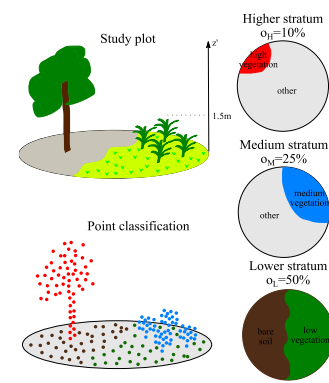

**Data normalization.** The z-value of each point is normalized by subtracting the z-value of the lowest point in a 0.5 m-cylindrical neighborhood. Moreover, all values are normalized between 0 and 1 over the entire dataset.

**Annotations.** Each plot has been annotated by a human expert *in situ* providing the lower, medium, and higher stratum occupancy ratio, $(ô_L, ô_M, ô_H) \in [0,1]^3$ respectively (Figure 1). The occupancy value $ô_L$ characterizes the proportion of the ground surface occupied by grass or low vegetation, as opposed to rocks, soil, or sand. $ô_M$ characterizes the proportion of the surface of the plot occupied by the footprint of medium vegetation located between 0.5 and 1.5 m. This type of vegetation, typically bush-like, is the most accessible by pasture animals. Finally, the higher stratum occupancy $ô_H$ is defined as the ratio of the plot surface occupied by the footprint of the canopy of trees over 1.5 m. Note that the trunks of trees over 1.5 m do not contribute to the medium occupancy. We argue that regressing the occupancy maps is intrinsically a semantic-constrained endeavor, and we thus opt for a neural network-based machine learning method.

Figure 1: **Objective.** Our method aims to predict the vegetation occupancy of three strata from a point cloud.

## 2.2 Methodology

Given the attributed 3D point cloud corresponding to a study plot (see Figure 1), we aim to produce the vegetation occupancy maps for each vegetation stratum (lower, medium and higher level). We first compute a soft prediction for each point among four different classes: bare soil, low vegetation, medium vegetation, and high vegetation. We then project the resulting probabilistic point prediction onto the rasterized disks corresponding to the three target strata. The occupancy ratio at plot level is then obtained by averaging the prediction in each stratum. We propose a weakly-supervised scheme which allows the network to predict a class for each 3D point as well as vegetation occupancy rasters while only using aggregated occupancies, corresponding to 3 values for each plot.

**Pointwise prediction.** Given the relative geometric simplicity of single plots, we use an architecture inspired by the PointNet network of Charles et al. (2017) to compute the pointwise predictions. We denote the predicted probabilities for a point $i \in [1, 2, ..., N]$ as follows: $(p_{i,S})$ for bare soil, $(p_{i,L})$ for lower, $(p_{i,M})$ for medium and $(p_{i,H})$ for high vegetation respectively (Figure 2).

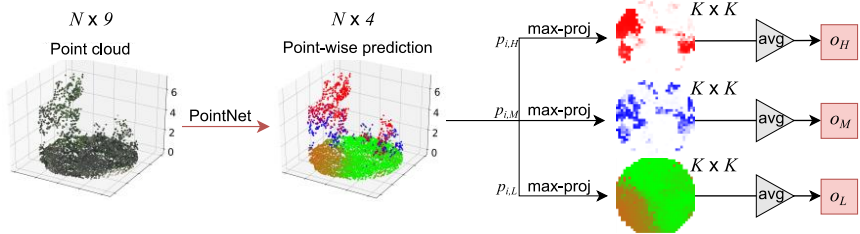

Figure 2: **Neural Architecture.** Our network performs the semantic segmentation of a 3D point cloud within four different classes. The resulting probabilities are projected onto rasters corresponding to different strata. Finally, the occupancies map are aggregated into the stratum vegetation ratio.

**Point projection.** The pointwise predictions are used to compute occupancy maps for each stratum, as shown in Figure 2. We consider three rasters of $K \times K$ pixels aligned with the projection of the cylindrical plot on the horizontal axes, and corresponding to the lower, medium, and higher vegetation strata. We associate the pixel $j$ of stratum $s \in \{L, M, H\}$ with the set of 3D points $\Pi_j$ whose vertical projection falls in the pixel's extent. We define the pixel occupancy $o_s^j$ as the maximum for all points of $\Pi_j$ of the probability of belonging to the vegetation stratum $s$:

$$o_s^j = \max_{i \in \Pi_j} p_{i,s} \ . \tag{1}$$

Finally, the stratum occupancy ratios $o_L, o_M, o_H$ are defined as the average of the pixels' occupancies in the corresponding stratum:

$$o_s = \frac{1}{K^2} \sum_{j \in K \times K} o_s^j \ . \tag{2}$$

**Elevation modeling.** The model described above does not explicitly model the distribution of elevations within each stratum. In theory, points that are several meters above the ground can contribute to the lower stratum as long as the stratum-wise aggregated values are in agreement with the ground truth. We propose to explicitly model the elevation of points within each stratum in an automated way with the goal of making the occupancy maps more interpretable, and increasing generalizability.

We model the normalized elevation of all points of all clouds with a mixture of two Gamma distributions corresponding to the lower stratum, and to the medium and higher strata respectively. The distribution parameters can be efficiently estimated with the expectation–conditional–maximization algorithm, as detailed by Young et al. (2019). Using the Bayes theorem, this allows us to compute the likelihood of the elevation of each point given its stratum prediction.

**Loss functions.** The model is trained using three loss functions: (i) the mean absolute error between the predicted and ground truth plot occupancy, (ii) the average entropy of each pixel occupancy value, (iii) the average negative log-likelihood of points' elevation conditioned by their pointwise predictions:

$$l = l_{data} + \alpha l_{entropy} + \lambda l_{likelihood} \tag{3}$$

with *α* and *λ* regularization strengths, set to *0.2* and *1* respectively during all experiments. We chose to regularize with the entropy of pixel occupancy to implement the prior that most pixels should be either empty (no vegetation) or full (completely covered).

Our network is implemented in PyTorch and trained with ADAM optimizer and a batch size of 20 and with raster size $K = 32$. Our code, the data, as well as the precise configuration of all layers can be accessed at the following URL: https://github.com/ekalinicheva/plot_vegetation_coverage.

## 3. Results

To assess the performance of our model, we implemented a handcrafted approach and a simple deep learning baseline. The handcrafted approach classifies points among the different strata with a decision tree algorithm based on colors and elevation, while the rest of our pipeline is unchanged. The deep learning baseline directly predicts the stratum occupancy: $o_L, o_M, o_H$ from the raw 3D point cloud with a simple PointNet network.

The qualitative results of our method and concurrent approaches are presented in Table 1. Our method outperforms the baselines, at the cost of added computation time compared to the simple deep baseline. Moreover, as seen in Figure 3, our method also allows visualizing the stratum occupancies.

Table 1. **Quantitative Results.** We report the accuracy of the predicted aggregated plot occupancy, along with the inference speed.

| Method | Absolute error, % | | | | Inference time, plots/s |
|---|---|---|---|---|---|
| stratum | lower | medium | higher | average | |
| Handcrafted | 21.9 | 20.7 | 10.3 | 17.6 | 20 |
| PointNet Baseline | 17.4 | 13.5 | 7.7 | 12.8 | 400 |
| Ours | 15.5 | 13.6 | 7.5 | 12.2 | 125 |

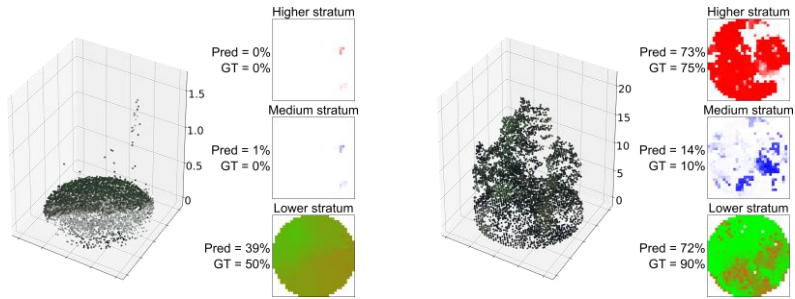

Figure 3: **Qualitative Results.** Our method predicts aggregated stratum occupancy along with the corresponding rasterized occupancy maps. Here, the pixels are colored according to the value of the predicted occupancy: shades of green, blue, and red indicate pixels with high-predicted vegetation coverage for the lower, medium, and higher strata respectively, while brown corresponds to bare soil.

## 4. Conclusion

In this paper, we presented a 3D deep learning method for predicting occupancy across vegetation strata. Using only three aggregated values per example plot, our model is able to perform a pointwise classification and to produce vegetation occupancy rasters with a high precision and a small computational cost. We also release the first deep learning-dataset to train and evaluate such methods.

## Acknowledgments

This study has been co-funded by CNES (TOSCA FRISBEE Project, convention n°200769/00) and CONFETTI Project (Nouvelle Aquitaine Region project, France).